# Causal Robot Communication Inspired by Observational Learning Insights

Zhao Han,[1*] Boyoung Kim,[2] Holly A. Yanco,[3] Tom Williams[1]

[1]Colorado School of Mines, MIRRORLab, 1500 Illinois St., Golden, CO 80401, USA
[2]George Mason University, ALPHAs Lab, 4400 University Dr., Fairfax, VA 22030, USA
[3]University of Massachusetts Lowell, Human-Robot Interaction Lab, 1 University Ave., Lowell, MA 01854, USA
{zhaohan, twilliams}@mines.edu, bkim55@gmu.edu, holly@cs.uml.edu

**Abstract**

Autonomous robots must communicate about their decisions to gain trust and acceptance. When doing so, robots must determine which actions are *causal*, i.e., which directly give rise to the desired outcome, so that these actions can be included in explanations. In behavior learning in psychology, this sort of reasoning during an action sequence has been studied extensively in the context of imitation learning. And yet, these techniques and empirical insights are rarely applied to human-robot interaction (HRI). In this work, we discuss the relevance of behavior learning insights for robot intent communication, and present the first application of these insights for a robot to efficiently communicate its intent by selectively explaining the causal actions in an action sequence.

## 1 Introduction

Interactive robots need to be able to communicate effectively about their intended actions in order to be trusted and accepted. For example, robots often need to be able to proactively communicate about their intended behaviors and the rationale for those behaviors (Peng et al. 2019; Baraglia et al. 2016). In this paper, we explore the insights for this process that might be drawn from the psychological subfield of behavior learning, concerned with "the acquisition of attitudes, values and styles of thinking and behaving through observation of the examples provided by others (i.e., models)" (Bandura 2008). Of particular relevance within behavior learning is *imitation learning*, studying how children and adults (Whiten et al. 2016; Buttelmann et al. 2017) learn to imitate *action sequences* of others (Buchsbaum et al. 2011).

Psychological research on imitation learning has uncovered two types of behavioral patterns in people's imitation learning. Some researchers found that, instead of copying exactly the same sequence of actions demonstrated by others, people can *rationally* learn to extract and produce most causal actions necessary to achieve some outcome (Buttelmann et al. 2017). Others found that people tend to *over-imitate*, copying not only necessary and causal behaviors but also unnecessary and non-causal behaviors, because they believe those behaviors are normative (Kenward 2012). Many robotics researchers have used imitation learning theories to model how robots can learn from demonstration; but we argue that the rational imitation paradigm should also be used to understand how robots can generate explanations that are understandable to human observers.

Similarly, while roboticists have a long history of studying the structure and use of action sequences (Haazebroek, Van Dantzig, and Hommel 2011; Nakawala et al. 2018), including their representation as state machines or behavior trees (Colledanchise and Ögren 2018), these efforts typically focused on underlying representation structures that robots can use to represent behaviors (Haazebroek, Van Dantzig, and Hommel 2011). In contrast, we argue that psychological insights regarding action sequence representation from human's perspectives should also be used to understand how robots can generate effective explanations regarding their own internal states (Han et al. 2021; Brooks et al. 2010).

If robots leverage structured representations of action sequences to communicate their intents, a key question is what elements of those sequences to communicate. Researchers in robotics, HRI, and explainable AI have used a variety of approaches to choose which states and actions to communicate. For example, Huang et al. (2018) present an entropy-based method for identifying critical states in the context of autonomous driving, and Olson et al. (2019) formulate an optimization problem to generate *counterfactual* states in the context of an Atari game. Yet technical research in this area has left the highly relevant psychological literature on observational learning (including its consideration of causality) relatively underexplored. For example, this literature suggests that humans differentiate between causal and non-causal actions in their imitations. We argue that robots must similarly differentiate between causal and non-causal actions in their explanations, as inclusion of all low-level actions would be overly verbose. Decades of previous work have demonstrated the importance of brevity (i.e., following Grice's maxims of quantity and manner (Grice 1975)) in automated explanation generation (Young 1999; Singh et al. 2021), and demonstrated human preferences for (and higher performance under) brevity (Bohus and Rudnicky 2008) with both dialogue system and robots (Han, Phillips, and Yanco 2021). In this paper, we thus draw on additional

[*]Portions of this work was done while Zhao Han was with UMass Lowell. This work was supported in part by the Office of Naval Research (N00014-18-1-2503) and NSF (IIS-1909864).
The AAAI SSS-22 Symposium "Closing the Assessment Loop: Communicating Proficiency and Intent in Human-Robot Teaming"

insights from the field of observational learning to identify how robots can choose the most salient causal actions to communicate in their explanations.

## 2 Applying Observational Learning Insights

A key theory from observational learning that can be applied is **social cognitive learning theory (SCLT)**, which seeks to explain how humans observe and learn the sequences of actions others use to achieve desired outcomes (Bandura 2008). SCLT considers four subprocesses of observation: *(1) attention*: Observers must attend to and perceive the modeling episode to profit from guidance (Yussen 1974); *(2) retention*: Observers must discriminate and symbolically represent the modeled behavior to make it easy to recall (Bandura and Barab 1971); *(3) motor skills*: Observers need key motor skills to translate knowledge into physical movements; *(4) motivation*: Observers must be motivated to minimize departure of learned actions from demonstrated actions.

In a situation where a robot attempts to communicate the intent underlying its action sequence, the robot can be viewed as "a model" and the human as "an observer" in SCLT. In such a situation, the four subprocesses above have important implications for how a robot should convey its intent to humans.

First, we found that people prefer robots to get their *attention* by addressing them before verbally explaining (Han, Phillips, and Yanco 2021). Previously, Bruce, Nourbakhsh, and Simmons (2002) found attention is key to robots' success at initiating communication. For *retention*, robots must provide humans an easily memorable summary of their internal states (Brooks et al. 2010) to facilitate accurate symbolic representation of those states in the mind of their interlocutors. This may be achieved by organizing and selecting high-level behaviors first (Han et al. 2021) rather than directly externalizing arbitrary internal states. For *motor skills*, non-human-like *and* non-deterministic robotic movements, produced by probabilistic methods (LaValle et al. 1998; Thrun, Burgard, and Fox 2005), might be troublesome to understand and may not lead to quality physical responses. The final component of Bandura's social cognitive learning theory is ensuring observer motivation. However, in the case of robot explanation generation, we can assume that the listener who requested an explanation is motivated to listen to and try to understand that explanation, under an assumption of good faith.

This literature also provides insights as to how demonstrators can best model behavior. Researchers have explored what leads people to imitate (Hilbrink et al. 2013). Brugger et al. (2007) showed that infants only copied causal action sequences, where the first action is causally necessary for the following action, ignoring adjacent, non-causal, unnecessary actions. Buttelmann et al. (2017) showed that imitators use intentionality to infer causal actions, and results show that intentional actions with markers like "there" and "here" are assumed to be purposive to understand causality.

Informed by the above insights from the observational learning literature, we have studied how a robot could communicate missing causal information in a versatile mobile manipulation task to convey its action intent. In that work, we considered a robot's need to explain previous actions in an environment that has changed since its actions were taken, with some objects had been replaced (Han and Yanco, under review), i.e., (1) where the robot grasped a misrecognized object, (2) where a ground obstacle led to a detour, and (3) where an object was placed into a wrong caddy section. In these scenarios, the robot cannot communicate the causes directly as it is not aware of perception failures.

We experimented with verbal and/or projection indicators with or without a replay of a robot's past physical actions. These indicators were inspired by the deictic indicators used to signal purposive intent as described above (Buttelmann et al. 2017). In the action sequence for the three subtasks, the robot selectively communicated *causal* actions that would lead to environment changes immediately after the actions.

Three verbal deictic expressions are spoken immediately before causal actions. To communicate where the misrecognized object was, the robot said: "Ok. I picked up a gearbox bottom from here," while its gripper was over the object and before lowering its gripper to grasp. For the detour, it said: "Ok. I didn't go straight to the caddy table because there was something on the floor in front of me on my left," before starting to drive itself. To inform where it placed, it said: "Ok. I placed the gearbox bottom into the near right section of the caddy," while its gripper was over the section of the caddy and before lowering and opening its gripper to release.

Leveraging the insights above from SCLT, "Ok" was used to draw *attention*; robot speech was selected to convey key, memorable episodes, to improve *retention* of the robot's actions and communications in humans' memory. Moreover, robot speech concerned high-level behaviors regarding nearby objects, rather than uninterpretable internal state information. Speech was timed to avoid talking during non-human-like arm movements that would be hard to *reproduce* even imaginarily.

Further inspired by the use of deictic language to convey purposive intent, we also implemented three projection markers to indicate causality, similar to Williams et al. (2019a)'s *Mixed Reality Deictic Gestures* (cf. (Williams et al. 2018, 2019b; Hamilton, Tran, and Williams 2020)).

To evaluate our approach, we asked participants whether they inferred each missing causal information, confidence in their inferences, and how fast are their inferences. While results showed that combining physical replay with verbal and projection indicators best helped participants make these inferences, verbal deictic expressions alone best enabled users to infer robots' previous placement actions, while mixed reality deictic gestures (arrow projection markers) alone best enabled users to infer previous environmental state.

## 3 Conclusion

We applied observational learning insights to robot communication, and described our efforts toward observational learning informed explanation of missing causal information about a robot's past behaviors. This draws attention to under-explored dimensions of robot communication and explanation (cp.(Han, Phillips, and Yanco 2021; Hellström 2021)). Each subprocess of social cognitive learning theory represent a valuable direction for future work.

# References

Bandura, A. 2008. *Observational Learning*. American Cancer Society.

Bandura, A.; and Barab, P. G. 1971. Conditions governing nonreinforced imitation. *Developmental Psychology*, 5(2): 244.

Baraglia, J.; Cakmak, M.; Nagai, Y.; Rao, R.; and Asada, M. 2016. Initiative in robot assistance during collaborative task execution. In *2016 11th ACM/IEEE International Conference on Human-Robot Interaction (HRI)*, 67–74. IEEE.

Bohus, D.; and Rudnicky, A. I. 2008. Sorry, I Didn't Catch That! In *Recent trends in discourse and dialogue*, 123–154. Springer.

Brooks, D.; Shultz, A.; Desai, M.; Kovac, P.; and Yanco, H. A. 2010. Towards state summarization for autonomous robots. In *Symposium on Dialog with Robots at 2010 AAAI Fall Symposium Series*.

Bruce, A.; Nourbakhsh, I.; and Simmons, R. 2002. The role of expressiveness and attention in human-robot interaction. In *Proceedings 2002 IEEE international conference on robotics and automation (Cat. No. 02CH37292)*, volume 4, 4138–4142. IEEE.

Brugger, A.; Lariviere, L. A.; Mumme, D. L.; and Bushnell, E. W. 2007. Doing the right thing: Infants' selection of actions to imitate from observed event sequences. *Child development*, 78(3): 806–824.

Buchsbaum, D.; Gopnik, A.; Griffiths, T. L.; and Shafto, P. 2011. Children's imitation of causal action sequences is influenced by statistical and pedagogical evidence. *Cognition*, 120(3): 331–340.

Buttelmann, D.; Schieler, A.; Wetzel, N.; and Widmann, A. 2017. Infants' and adults' looking behavior does not indicate perceptual distraction for constrained modelled actions – An eye-tracking study. *Infant Behavior and Development*, 47: 103–111.

Colledanchise, M.; and Ögren, P. 2018. *Behavior trees in robotics and AI: An introduction*. CRC Press.

Grice, H. P. 1975. Logic and conversation. In *Speech acts*, 41–58. Brill.

Haazebroek, P.; Van Dantzig, S.; and Hommel, B. 2011. A computational model of perception and action for cognitive robotics. *Cognitive processing*, 12(4): 355.

Hamilton, J.; Tran, N.; and Williams, T. 2020. Tradeoffs Between Effectiveness and Social Perception When Using Mixed Reality to Supplement Gesturally Limited Robots. In *3rd International Workshop on Virtual, Augmented, and Mixed Reality for Human-Robot Interaction (VAM-HRI)*.

Han, Z.; Giger, D.; Allspaw, J.; Lee, M. S.; Admoni, H.; and Yanco, H. A. 2021. Building the Foundation of Robot Explanation Generation Using Behavior Trees. *ACM Transactions on Human-Robot Interaction (THRI)*, 10(3).

Han, Z.; Phillips, E.; and Yanco, H. A. 2021. The Need for Verbal Robot Explanations and How People Would Like a Robot to Explain Itself. *ACM Transactions on Human-Robot Interaction (THRI)*, 10(4): 1–42.

Han, Z.; and Yanco, H. A. Under review as of Mar. 2022. Communicating Missing Causal Information to Explain a Robot's Past Behavior. *ACM Transactions on Human-Robot Interaction (THRI)*.

Hellström, T. 2021. The relevance of causation in robotics: A review, categorization, and analysis. *Paladyn, Journal of Behavioral Robotics*, 12(1): 238–255.

Hilbrink, E. E.; Sakkalou, E.; Ellis-Davies, K.; Fowler, N. C.; and Gattis, M. 2013. Selective and faithful imitation at 12 and 15 months. *Developmental Science*, 16(6): 828–840.

Huang, S. H.; Bhatia, K.; Abbeel, P.; and Dragan, A. D. 2018. Establishing appropriate trust via critical states. In *2018 IEEE/RSJ International Conference on Intelligent Robots and Systems (IROS)*, 3929–3936. IEEE.

Kenward, B. 2012. Over-imitating preschoolers believe unnecessary actions are normative and enforce their performance by a third party. *Journal of experimental child psychology*, 112(2): 195–207.

LaValle, S. M.; et al. 1998. Rapidly-exploring random trees: A new tool for path planning. Technical report, Iowa State University.

Nakawala, H.; Goncalves, P. J.; Fiorini, P.; Ferringo, G.; and De Momi, E. 2018. Approaches for action sequence representation in robotics: a review. In *2018 IEEE/RSJ International Conference on Intelligent Robots and Systems (IROS)*, 5666–5671. IEEE.

Olson, M. L.; Neal, L.; Li, F.; and Wong, W.-K. 2019. Counterfactual states for atari agents via generative deep learning. *IJCAI 2019 Workshop on Explainable Artificial Intelligence*.

Peng, Z.; Kwon, Y.; Lu, J.; Wu, Z.; and Ma, X. 2019. Design and evaluation of service robot's proactivity in decision-making support process. In *Proceedings of the 2019 CHI Conference on Human Factors in Computing Systems*, 1–13.

Singh, A. K.; Baranwal, N.; Richter, K.-F.; Hellström, T.; and Bensch, S. 2021. Verbal explanations by collaborating robot teams. *Paladyn, Journal of Behavioral Robotics*, 12(1): 47–57.

Thrun, S.; Burgard, W.; and Fox, D. 2005. *Probabilistic robotics/*. MIT Press.

Whiten, A.; Allan, G.; Devlin, S.; Kseib, N.; Raw, N.; and McGuigan, N. 2016. Social learning in the real-world: 'Over-imitation' occurs in both children and adults unaware of participation in an experiment and independently of social interaction. *PloS one*, 11(7): e0159920.

Williams, T.; Bussing, M.; Cabrol, S.; Boyle, E.; and Tran, N. 2019a. Mixed Reality Deictic Gesture for Multi-Modal Robot Communication. In *Proceedings of the 14th ACM/IEEE International Conference on Human-Robot Interaction*.

Williams, T.; Bussing, M.; Cabrol, S.; Lau, I.; Boyle, E.; and Tran, N. 2019b. Investigating the Potential Effectiveness of Allocentric Mixed Reality Deictic Gesture. In *International Conference on Virtual, Augmented, and Mixed Reality*.

Williams, T.; Tran, N.; Rands, J.; and Dantam, N. T. 2018. Augmented, Mixed, and Virtual Reality Enabling of Robot

Deixis. In *International Conference on Virtual, Augmented, and Mixed Reality (VAMR)*.

Young, R. M. 1999. Using Grice's maxim of quantity to select the content of plan descriptions. *Artificial Intelligence*, 115(2): 215–256.

Yussen, S. R. 1974. Determinants of visual attention and recall in observational learning by preschoolers and second graders. *Developmental Psychology*, 10(1): 93.